\documentclass{article}

\usepackage[dblblindworkshop,final]{neurips_2025}
\workshoptitle{Rethinking AI: Efficiency, Frugality, and Sustainability}

\usepackage[utf8]{inputenc} 
\usepackage[T1]{fontenc}    
\usepackage{hyperref}       
\usepackage{url}            
\usepackage{booktabs}       
\usepackage{amsfonts}       
\usepackage{nicefrac}       
\usepackage{microtype}      
\usepackage{xcolor}         
\usepackage{algorithm}
\usepackage{algorithmic}
\usepackage{pifont}
\usepackage{todonotes}
\usepackage{paralist}
\usepackage{subcaption}
\usepackage{tikz}
\usetikzlibrary{arrows.meta,positioning,shapes.geometric,shadows.blur,fit}

\usepackage{colortbl}

\usepackage{amsmath, amssymb}
\usepackage{cleveref}

\title{Quant-Trim in Practice: Improved Cross-Platform Low-Bit Deployment on Edge NPUs}

\author{%
  Rayen Dhahri \quad Steffen Urban \\
  Corporate Research and Technology, Carl Zeiss AG\\
  \texttt{\{rayen.dhahri, steffen.urban\}@zeiss.com}
}

\newcommand{\nanosamrepo}{\url{https://github.com/NVIDIA-AI-IOT/nanosam}}

\newcommand{\hailo}{Hardware A}
\newcommand{\mx}{Hardware B}
\newcommand{\rknn}{Hardware C}
\newcommand{\sakura}{Hardware D}

\newcommand{\WA}[2]{\textsf{W#1/A#2}}     
\newcommand{\stat}{\textsc{static}}        
\newcommand{\qat}{\textsc{qat}}            
\newcommand{\na}{\textsc{n/a}}
\newcommand{\yes}{\textcolor{green!60!black}{Yes}}
\newcommand{\no}{\textcolor{red!70!black}{No}}
\newcommand{\cond}{\textcolor{blue!70!black}{Cond.}}
\newcommand{\cmark}{\ding{51}}
\newcommand{\xmark}{\ding{55}}
\begin{document}

\maketitle
\begin{abstract}

Specialized edge accelerators rely on low-bit quantization, but vendor compilers differ in scaling, clipping, and kernel support, often as black boxes. The same floating-point (FP) checkpoint can therefore yield inconsistent accuracy across backends, forcing practitioners to tweak flags or refactor models to vendor-friendly operator subsets. We introduce \emph{Quant-Trim}, a training-phase method that produces a hardware-neutral checkpoint robust to backend and precision choices. It combines \emph{progressive fake quantization} to align training with the deployed integer grid and \emph{reverse pruning} to tame outlier-driven scale inflation while preserving learnability. Quant-Trim is agnostic to quantization schemes (symmetric/asymmetric, per-tensor/per-channel, INT8/INT4) and requires no vendor-specific graph changes. Across models and tasks, it narrows the FP→low-bit gap, reduces dependence on compiler heuristics/calibration, and avoids per-backend retraining. We report accuracy and edge metrics latency, throughput, energy/inference, and cost under static/dynamic activation scaling and varying operator coverage.

\end{abstract}

\section{Introduction}

Deploying neural networks in production typically imposes high demands on latency, power consumption, and cost, especially on the edge. Quantization is a primary lever: mapping floating-point parameters and activations to low-bit integers reduces compute and bandwidth without changing architecture~\citep{gholami2021survey,krishnamoorthi2018quantization,jacob2018quantization}. In practice, however, low-bit accuracy is dominated by \emph{activations}: their wide dynamic range, nonstationary statistics, and outliers make them fragile under clipping and coarse scaling~\citep{choi2018pact,esser2019lsq,xiao2022smoothquant,lin2023awq,dettmers2022llmint8}.

A common compromise is to keep activations in FP16/BF16 while quantizing weights, which improves stability but retains higher memory traffic and power consumption~\citep{micikevicius2018mixed,kalamkar2019bfloat16}. By contrast, many NPUs/ASICs enforce INT8 for both weights and activations with \emph{offline} calibration~\citep{jacob2018quantization,tflite_ptq,tensorrt_int8,openvino_ptq}. This heterogeneity means that the same FP checkpoint can produce \emph{widely varying} accuracy across backends due to opaque scaling/fusion choices; and the need to redesign architectures for “quantization-friendliness,” distill into restricted operator sets, or retrain per target~\citep{gholami2021survey,esser2019lsq,choi2018pact,nagel2020adaround}. Methods that rescale activations into weights~\citep{xiao2022smoothquant} or isolate outliers in higher precision~\citep{dettmers2022llmint8} help, but typically assume backend support that is not universal on NPUs and remain restricted under vendor-specific quantizers aligning with their compilers.

We introduce \emph{Quant-Trim}, a training-time procedure that produces a single, hardware-agnostic checkpoint robust to vendor compilers and precision regimes. Quant-Trim couples (i) \emph{progressive fake quantization}, which interpolates between FP32 and low-bit scales to align forward statistics smoothly with the deployment grid, and (ii) \emph{reverse pruning}, which pins extreme weights at the quantization boundary to prevent scale inflation while preserving learnability. By aligning train-time numerics with deployment, our recipe reduces sensitivity to backend scaling/clipping, enabling reliable INT8 export without per-backend retraining.

\paragraph{Contributions.} 
\begin{itemize} 
\item \textbf{Robust Training with Fake Quantization.} We introduce a progressive fake-quantization curriculum that minimizes activation-induced errors across various deployment precisions, preventing optimization collapse.
\item \textbf{Scale Management through Reverse Pruning.} Our pin-at-boundary method effectively mitigates weight outlier influence while preserving gradient flow and model capacity. 
\item \textbf{Edge Efficiency Evaluation.} We benchmark multiple edge devices (GPUs/SoCs/NPUs) and quantify latency, power, and energy-per-inference, highlighting their promise for efficient and greener AI deployment.
\end{itemize}

\section{Background}\label{sec:background}

Efficient deployment at the edge relies on numeric precision. Floating-point formats (FP32, FP16, BF16) trade accuracy for bandwidth and energy, enabling scalable training and inference~\citep{micikevicius2018mixed}. Integer quantization (e.g., INT8/INT4) goes further by mapping tensors to fixed grids; with scale $s$ and zero-point $z$,
\[
Q(x)=\operatorname{clip}\!\Big(\big\lfloor x/s \big\rceil + z,\; q_{\min}, q_{\max}\Big),
\qquad
\hat{x}=s\,(Q(x)-z),
\]
typically using symmetric ($z{=}0$) weights and asymmetric activations~\citep{jacob2018quantization,krishnamoorthi2018quantization,gholami2022survey}. Per-channel scales often improve accuracy for conv/linear layers, but support varies by compiler.

Different hardware vendors ship black-box compilation and optimization algorithms tailored for their hardware and hence come with varying constraints, e.g.: 
\begin{itemize}
\item Per-channel / asymmetric kernels
\item Mixed precision (tunable per layer)
\item Operator support
\item Post-training layer fusion 
\item Histogram observers
\end{itemize}
The hardware used in our evaluations is covered under \cref{sub-sec:hardplatform} and \cref{subsec:form-factors}. Many NPUs require static INT8 for both weights and activations; GPUs allow mixed regimes (e.g., W8/A16) and emerging low-FP formats (FP8)~\citep{kim2023fp8}. These differences mean the \emph{same} FP checkpoint can yield divergent low-bit accuracy across backends.

We seek a training-time procedure that produces a \emph{single} checkpoint whose low-bit behavior is stable under heterogeneous compiler choices (scaling, clipping, kernel availability), without backend-specific graph edits. We base our approach on fake quantization (to align train-time forwards with deployed integer numerics) and scale control via reverse pruning by compressing the range.

\section{Methodology}\label{sec:quanttrim}

\begin{figure}
    \centering
    \includegraphics[width=\linewidth]{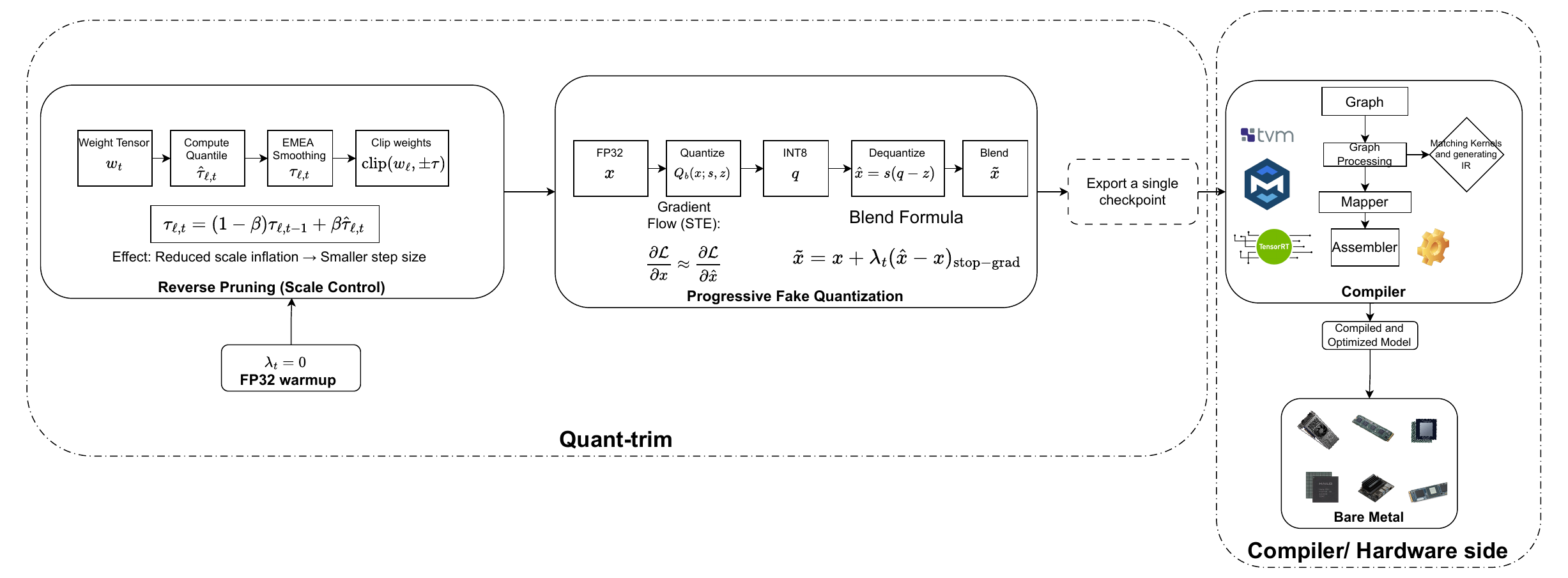}
    \caption{\textbf{Quant-Trim training pipeline.} Our method combines two key components: (1) \textbf{Reverse Pruning} clips extreme weights at robust quantile thresholds $\tau_{\ell,t}$ to prevent scale inflation while retaining representational power, and (2) \textbf{Progressive Fake Quantization} smoothly interpolates between FP32 and INT8 execution via a curriculum schedule $\lambda_t$ to avoid optimization collapse. The blend coefficient gradually increases from 0 (full FP32 warmup) through a quartic ramp to 1 (full fake quantization), while computing per-tensor/channel scales and zero-points. Gradients flow via straight-through estimator (STE). The final model exports to standard ONNX without custom operators, ensuring compatibility with NPU compilers.}
    \label{fig:quanttrim-fancy}
\end{figure}

In this section we introduce our \textbf{Quant-Trim} approach. It combines two key components. First is the \emph{progressive fake quantization} that smoothly interpolates between FP32 and low-bit execution to avoid optimization collapse, and the \emph{reverse pruning} step that pins extreme weights at the quantization boundary to prevent a few values from inflating the scale while retaining representational power.
The Quant-Trim training workflow is depicted in \cref{fig:quanttrim-fancy}.
In the following subsection we describe each step in detail.
Methods and earlier work that motivated our work are described in \cref{sec:Related}.


\subsection{Problem definition and notation}

\subsubsection{Uniform quantizer and STE.}
For a tensor $x$ and bit-width $b=8$, we use a uniform fake quantizer with scale $s>0$ and zero-point $z$:
\[
Q_{b}(x;s,z) \;=\; \operatorname{clip}\!\Big(\big\lfloor \tfrac{x}{s} + z \big\rceil,\; q_{\min},\; q_{\max}\Big),
\qquad
\hat{x}\;=\; s\big(Q_{b}(x;s,z)-z\big),
\]
where $(q_{\min},q_{\max})=(-128,127)$ for symmetric INT8 (weights) and $(0,255)$ for asymmetric UINT8 (activations).
We use the straight-through estimator (STE)~\cite{bengio2013estimating,hubara2017qnn,yin2019understandingstraightthroughestimatortraining}:
\[
\frac{\partial \mathcal{L}}{\partial x}\;\approx\;\frac{\partial \mathcal{L}}{\partial \hat{x}},
\]
and perform \emph{progressive blending} at each \emph{quantization point} (i.e., every weight tensor and designated activation location) with a strength $\lambda_t\in[0,1]$:
\[
\tilde{x} \;=\; x \;+\; \lambda_t\,(\hat{x}-x)_{\text{stop-grad}},
\]
We use a single global blend coefficient $\lambda_t$ shared across all quantization points at epoch $t$. Thus, $\lambda_t=0$ equals full FP32 precision and $\lambda_t=1$ is full fake-quant in forward. The gradients always follow FP32.

\subsubsection{Robust statistics and tensor quantiles}

\paragraph{Quantiles}
For a random variable $X$ with CDF $F_X$, the $p$-quantile is
\[
Q_X(p)\;:=\;F_X^{-1}(p)\;=\;\inf\{x:\,F_X(x)\ge p\},\quad p\in(0,1).
\]
Given samples $x_{1:n}$ with order statistics $x_{(1)}\!\le\!\cdots\!\le\!x_{(n)}$, the empirical quantile is
$\widehat Q_X(p)=x_{(\lceil pn\rceil)}$.
For large tensors we compute $\widehat Q$ on a random subsample $S_t$, $|S_t|\le S_{\max}$ (we use $S_{\max}=10^5$).
We write $p_{\mathrm{hi}},p_{\mathrm{lo}}\in(0,1)$ for upper/lower quantiles (e.g., $p_{\mathrm{hi}}=0.999$, $p_{\mathrm{lo}}=0.001$), $\mu\in(0,1)$ for EMA momentum (we use $\mu=10^{-3}$), and $\varepsilon>0$ (we use $10^{-6}$).

\paragraph{Per-tensor statistics} Let $b=8$. For \emph{weights} (symmetric), with $X=|w|$,
\[
m_t \;=\; Q_{|w|}(p_{\mathrm{hi}})\ \ (\text{or }\widehat Q^{(S)}),\qquad
\tilde m_t \;=\; (1-\mu)\,\tilde m_{t-1} + \mu\, m_t,
\]
\[
s_t^{(\mathrm{w})} \;=\; \frac{\max(\tilde m_t,\varepsilon)}{2^{b-1}-1},\qquad z^{(\mathrm{w})}=0.
\]
For \emph{activations} (asymmetric),
\[
a_t \;=\; Q_{x}(p_{\mathrm{lo}}),\quad b_t \;=\; Q_{x}(p_{\mathrm{hi}}),\qquad
\tilde a_t \;=\; (1-\mu)\,\tilde a_{t-1} + \mu\, a_t,\quad
\tilde b_t \;=\; (1-\mu)\,\tilde b_{t-1} + \mu\, b_t,
\]
\[
s_t^{(\mathrm{a})} \;=\; \frac{\max(\tilde b_t-\tilde a_t,\varepsilon)}{2^{b}-1},\qquad
z_t^{(\mathrm{a})} \;=\; \operatorname{clip}\!\Big(-\frac{\tilde a_t}{s_t^{(\mathrm{a})}},\,q_{\min},\,q_{\max}\Big).
\]
The same definitions apply per-output channel $c$ by replacing $w$ with $w_c$ (or $x$ with $x_c$) and computing $Q_{|w_c|}$ (or $Q_{x_c}$) along the channel axes; EMAs are then channel-wise.

\subsection{Reverse Pruning (Scale Control)}
Outlier weights inflate the effective scale. For layer $\ell$ with weights $w_\ell$, let
\[
\hat\tau_{\ell,t} \;=\; \widehat Q^{(S)}_{|w_\ell|}(p_{\mathrm{clip}}) \quad (\text{e.g., } p_{\mathrm{clip}}=0.95),
\qquad
\tau_{\ell,t} \;=\; (1-\beta)\,\tau_{\ell,t-1}+\beta\,\hat\tau_{\ell,t},
\]
with $\beta\in(0,1]$ an EMA momentum. Every $K$ epochs after warmup we \emph{pin}
\[
w_\ell \leftarrow \operatorname{clip}\big(w_\ell,\,-\tau_{\ell,t},\,\tau_{\ell,t}\big).
\]
For symmetric quantization the post-pruning step size satisfies
\[
\Delta' \;=\; \frac{\tau_{\ell,t}}{2^{b-1}-1} \;<\; \Delta \;=\; \frac{\max_i|w_{\ell,i}|}{2^{b-1}-1},
\]
allocating more representational levels to the bulk. Empirically, \cref{fig:dist-shift} shows compressed weight tails and narrower downstream activation ranges.

\subsection{Training Curriculum}
We schedule the \emph{global} blend $\lambda_t$ over epochs $t$ and apply it at every quantization point:
\[
\lambda_t \;=\;
\begin{cases}
0, & t < E_w \quad \text{(FP32 warmup)};\\[4pt]
\min\!\Big(0.5,\; \big(\tfrac{t-E_w}{E_f-E_w}\big)^{\!4}\cdot 0.5\Big), 
& E_w \le t < E_f \quad \text{(gentle quartic ramp)};\\[8pt]
0.5 + \Big(\min\!\big(1,\tfrac{t-E_f}{H}\big)\Big)^{\!2}\cdot 0.5, 
& t \ge E_f \quad \text{(quadratic to full)}.
\end{cases}
\]
Here $E_w$ is warmup end, $E_f$ is the end of the ramp, and $H$ is the horizon to reach $\lambda_t\!=\!1$.
At each quantization point we compute $\hat{x}$ using $Q_{b}(\cdot)$ with current $(s_t,z_t)$ from the robust statistics above and output $\tilde{x}=x+\lambda_t(\hat{x}-x)_{\text{stop-grad}}$; gradients use STE.

\subsection{Training Procedure and Export}
Putting it together:
\begin{compactenum}
\item \textbf{FP32 warmup} ($0{:}E_w$): train with $\lambda_t\!=\!0$.
\item \textbf{Reverse pruning onset} ($t\!=\!E_w$): start EMA thresholds $\tau_{\ell,t}$; pin every $K$ epochs.
\item \textbf{Progressive fake-quant} ($E_w{:}E_f$): enable fake-quant for weights (symmetric) and activations (asymmetric); update $(s_t,z_t)$ via robust EMA quantiles; blend with quartic $\lambda_t$.
\item \textbf{Advanced/final} ($t\ge E_f$): quadratic ramp to $\lambda_t\!=\!1$; keep STE for stability.
\end{compactenum}
We realize this via per-layer wrappers that (i) temporarily substitute quantized weights in forward while keeping FP32 master weights for updates, and (ii) insert activation fake-quant hooks after common nonlinearities. The final checkpoint is exported to ONNX and compiled with vendor toolchains (TensorRT/TVM/NPU compilers); the computational graph remains standard (no fused rescaling or non-standard formats).

\subsection{Mechanism and Intuition}
\begin{figure}[H]
    \centering
    \includegraphics[width=0.45\linewidth]{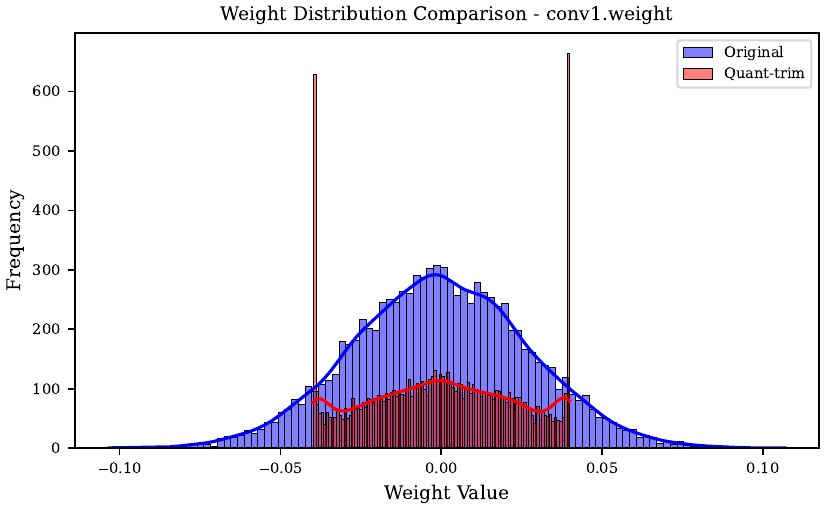}
    \includegraphics[width=0.45\linewidth]{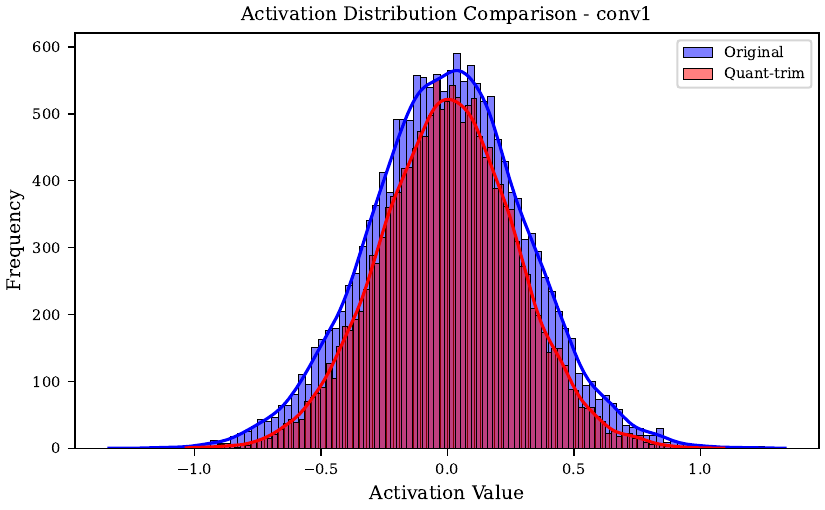}
    \caption{\textbf{Distributional effect of Quant-Trim.} Left: reverse pruning compresses weight tails, reducing scale inflation. Right: activations exhibit a narrower dynamic range, making INT8 mapping more stable.}
    \label{fig:dist-shift}
\end{figure}
\paragraph{Scale control.} Pinning the largest $p\%$ weights reduces the extreme order statistics that set the quantization scale, shrinking $\Delta$ and allocating more representational levels to the bulk.
\paragraph{Smooth noise injection.} The blend $\tilde{x}=x+\lambda_t(\hat{x}-x)_{\text{stop-grad}}$ gradually aligns train-time forward with deploy-time integer forward while the backward pass remains FP32-stable~\cite{bengio2013estimating,esser2019lsq}, reducing the distributional gap between FP32 and low-bit, as visualized in \cref{fig:dist-shift}.

\section{Related Work}\label{sec:Related}

\paragraph{QAT and learned quantizers.}
Quantization-aware training enhances low-bit deployment by learning clip ranges and step sizes (PACT, LSQ)~\citep{choi2018pact,esser2019lsq}. Extensions target transformers and mixed regimes, but accuracy can vary across compilers due to scale/zero-point handling and fusion rules~\citep{gholami2022survey}.

\paragraph{PTQ and calibration sensitivity.}
Post-training methods reduce retraining cost but remain sensitive to activation range estimation and small calibration sets (AdaRound, BRECQ)~\citep{nagel2020adaround,li2021brecqpushinglimitposttraining}. Weight-only routes help for LMs yet leave activation error unresolved (LLM.int8, AWQ, GPTQ)~\citep{dettmers2022llmint8,lin2023awq,frantar2023gptqaccurateposttrainingquantization}.

\paragraph{Activation outliers and distribution reshaping.}
A main failure mode is activation heavy tails under A8/W8 and A4/W4. Smoothing or reassigning scale to weights reduces clipping/rounding (SmoothQuant and variants)~\citep{xiao2022smoothquant}. Rotation/dual-transform and low-rank absorption further reduce tails for ultra-low bits~\citep{lin2024duquantdistributingoutliersdual,rotatedruntime2024,li2024svdquant}. Many rely on fused rescaling or side branches that are not universally supported on edge NPUs.

\paragraph{Pruning for scale control.}
Classical magnitude pruning targets sparsity, not scale robustness. Outlier-aware pruning prioritizes sensitive channels but is backend-specific~\citep{lee2024owq,yin2024owl}. Our \emph{reverse pruning} focuses on \emph{tail pinning}: clipping scale-setting weights to contract per-(tensor/channel) ranges before fake-quant—keeping the exported graph standard.

\paragraph{Backend heterogeneity.}
Edge compilers differ in per-channel/asymmetric support, placement of rescale ops, dynamic vs.\ static activation scaling, and allowed mixed-precision islands~\citep{jacob2018quantization,gholami2022survey}. This leads to cross-backend variance from the same FP checkpoint and motivates training-time robustness rather than backend-specific graph edits.

\paragraph{Regularization links.}
Bayesian/Laplace shrinkage can suppress heavy tails and improve sparsifiability; it is complementary to our scale-control view~\citep{immer2021scalable,laplace_redux_neurips21,dhahri2024spam}.

\section{Experiments}
We present large-scale optimization and experiments across multiple hardware platforms, tasks, and models to quantify NPU efficiency and the impact of varying numerical precision on devices that support multiple precisions. The experimental pipeline is detailed in \cref{sec:experm}. We also provide comprehensive hardware descriptions and specifications to contextualize the results under \cref{subsec:form-factors} and to make explicit the limitations of each platform and conduct an ablation study on the effect of Quant-Trim components in \cref{sec:ablation}. Device capabilities are summarized in \cref{tab:edge-npu-specs}.We then show the stability of the training results and the effect of our method on the end quantization results on the accelerators.

\begin{figure}[H]
    \centering
    \includegraphics[width=0.49\linewidth]{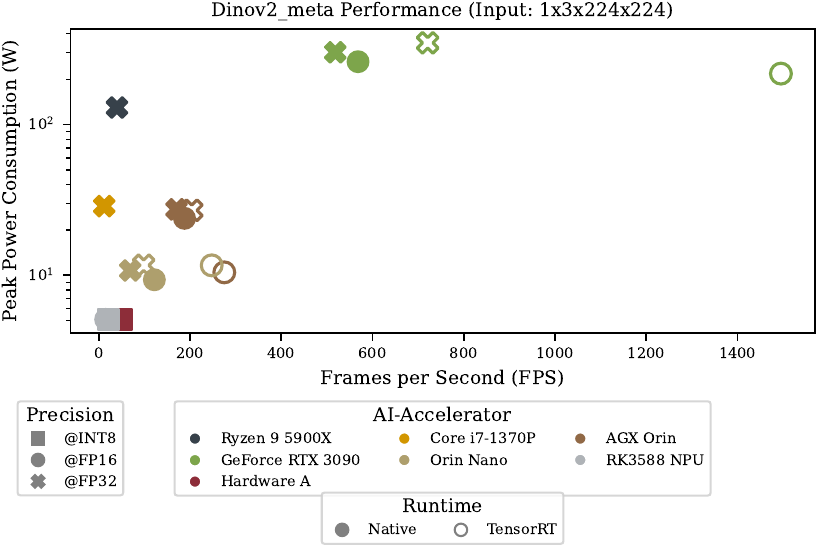}
    \includegraphics[width=0.49\linewidth]{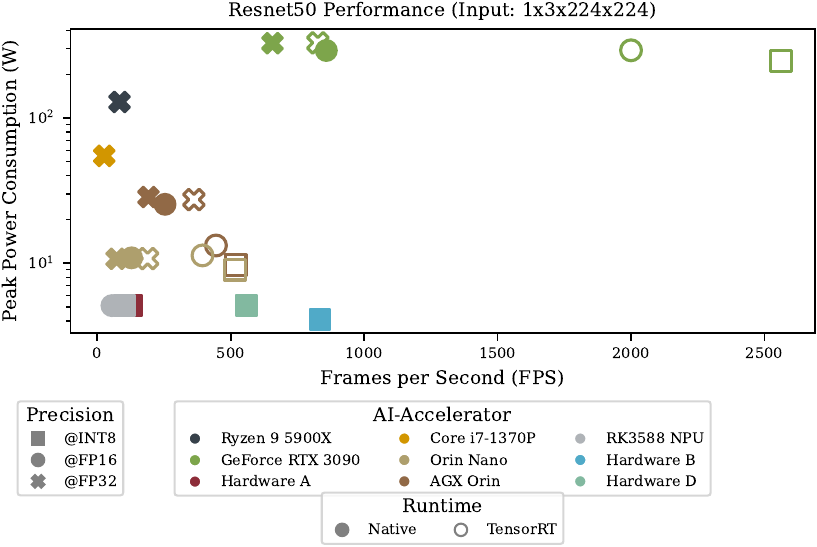}
    \caption{\textbf{Power–throughput trade-off for DINOv2 and ResNet-50.}
Batch{=}1, $224{\times}224$ input. 
$x$-axis: median FPS over 200 timed iters after 20 warm-ups; $y$-axis: average Peak-power (5 runs; whiskers show 5–95th percentiles).
\textbf{Encoding:} color = device; marker shape = precision; \emph{filled} markers = platform’s default runtime (NPUs: vendor runtime; NVIDIA: CUDA), \emph{unfilled} markers = TensorRT.
Left: DINOv2; Right: ResNet-50. Device specs in \cref{tab:edge-npu-specs}.}
    \label{fig:benchdinov}
\end{figure}

\begin{figure}[H]
\centering
\begin{subfigure}{0.48\linewidth}
    \centering
    \includegraphics[width=\linewidth]{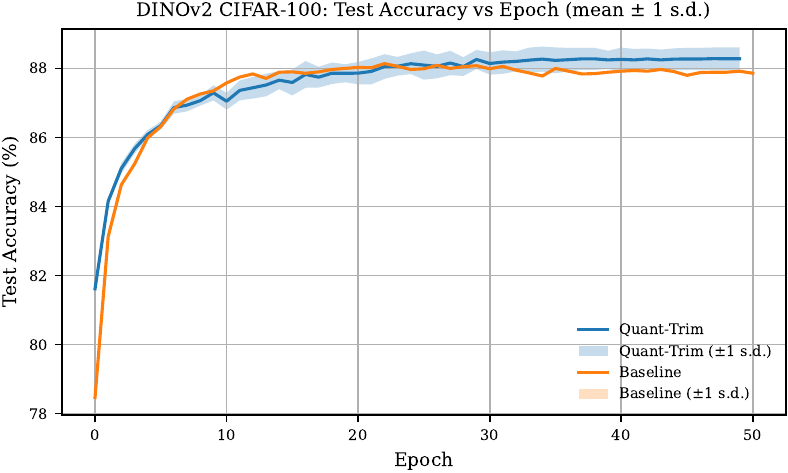}
    \caption{Top-1 accuracy (Quant-Trim vs.\ MAP).}
\end{subfigure}\hfill
\begin{subfigure}{0.48\linewidth}
    \centering
    \includegraphics[width=\linewidth]{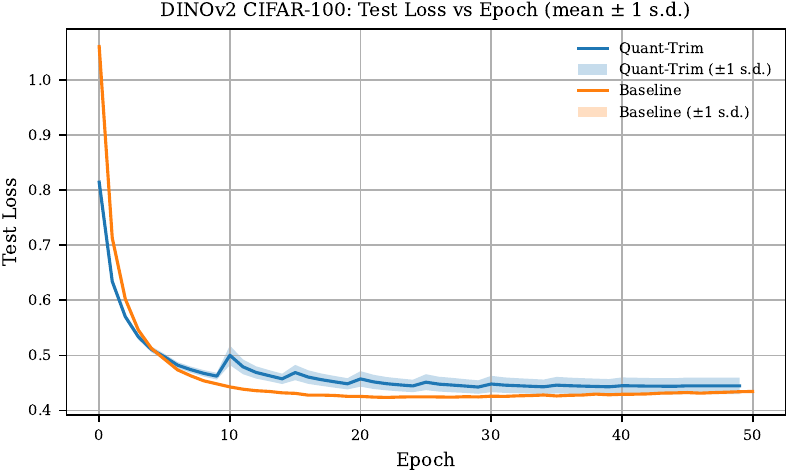}
    \caption{Test loss (Quant-Trim vs.\ MAP).}
\end{subfigure}
\caption{\textbf{Training dynamics on CIFAR-100 (DINOv2).} A small accuracy dip during the ramp phase is followed by convergence to MAP-level performance.}
\label{fig:DINOv2-dynamics}
\end{figure}
\subsection{Inference Results and Efficiency}
Within the same hardware, lower precision yields faster speeds. For NVIDIA GPUs and embedded Jetson devices, TensorRT gives a significant boost compared to using CUDA kernels. In general, for the shown models, the requirements to perform real-time tasks are slightly beyond 60 FPS to be able to build on top and take into consideration the latency that comes from data processing and sensor systems. We provide technical details about the system and the potential latency added for end deployment under \cref{sub-sec:hardplatform}.

\paragraph{Power Consumption vs Precision}
NPUs' energy consumption is very low in comparison to NVIDIA GPUs, which can pull up to 200~W, while NPUs do not exceed 10~W. Hardware that supports different precision like the Jetson family and GPU are 2 to 3 times faster at lower precision compared to FP32 for both ResNet-50 and DINOv2. The Power Consumption is measured on chip. 

\paragraph{Runtime Provider} Within the same hardware, we can see that for NVIDIA hardware (GPU and Jetson), TensorRT provides a large speedup for FP16, nearly tripling the speed of DINOv2 from 600 FPS to over 1500 FPS as shown in \cref{fig:benchdinov}.Additional results for U-Net \citep{ronneberger2015unet} and MobileNetV3-Small \citep{howard2019mobilenetv3} are shown in \cref{fig:mobilenet_unet_comparison}.

\subsection{Training dynamics}

\begin{figure}[H]
\centering
\includegraphics[width=0.8\linewidth]{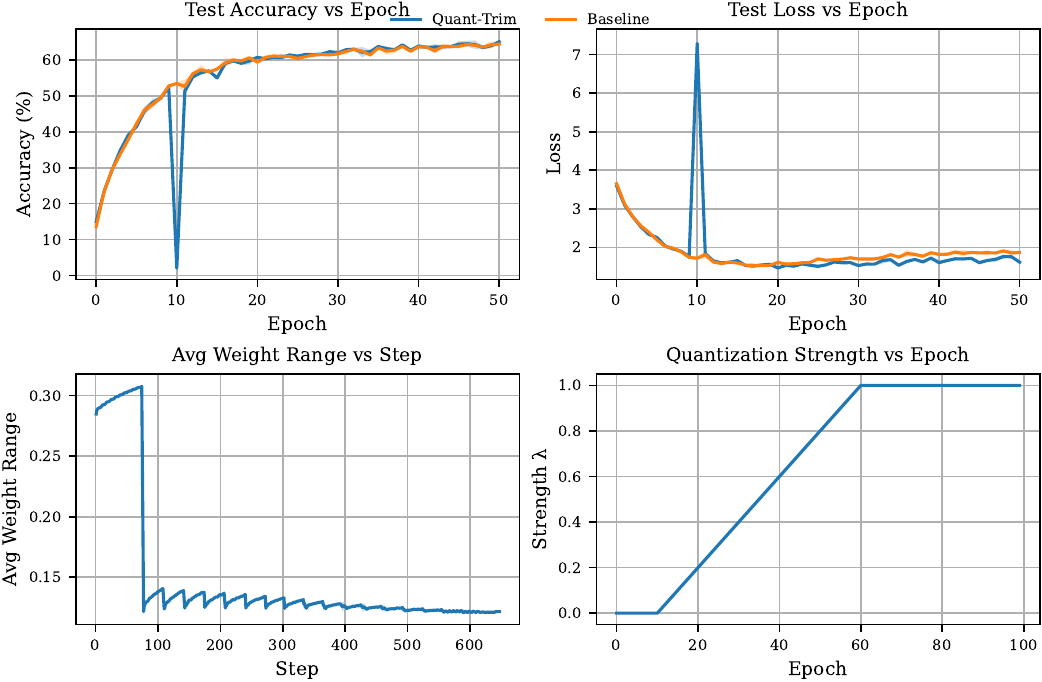} 
\caption{Quant-Trim exhibits a brief dip when fake-quantization ramps in, then recovers to near-baseline accuracy and loss by the end of training for ResNet on CIFAR-100.}
\label{fig:resnet-dynamics}
\end{figure}

\paragraph{Training dynamics \& convergence.} Similar to MAP training, Quant-Trim leads to full convergence at the end of training and similar predictive performance. Training stabilizes, ensuring progressive quantization with a slight drop once we start the fake-quantization process, but regains near complete accuracy by the end of training for ResNet-50 and DINOv2 as shown respectively in \cref{fig:resnet-dynamics} and \cref{fig:DINOv2-dynamics}. A similar trend is seen for ResNet-18 on Coco as shown in \cref{fig:resnet_coco}. 

\paragraph{Feature alignment (NanoSAM2).}

\begin{figure}[H]
  \centering
  \begin{subfigure}{0.48\linewidth}
    \centering
    \includegraphics[width=\linewidth]{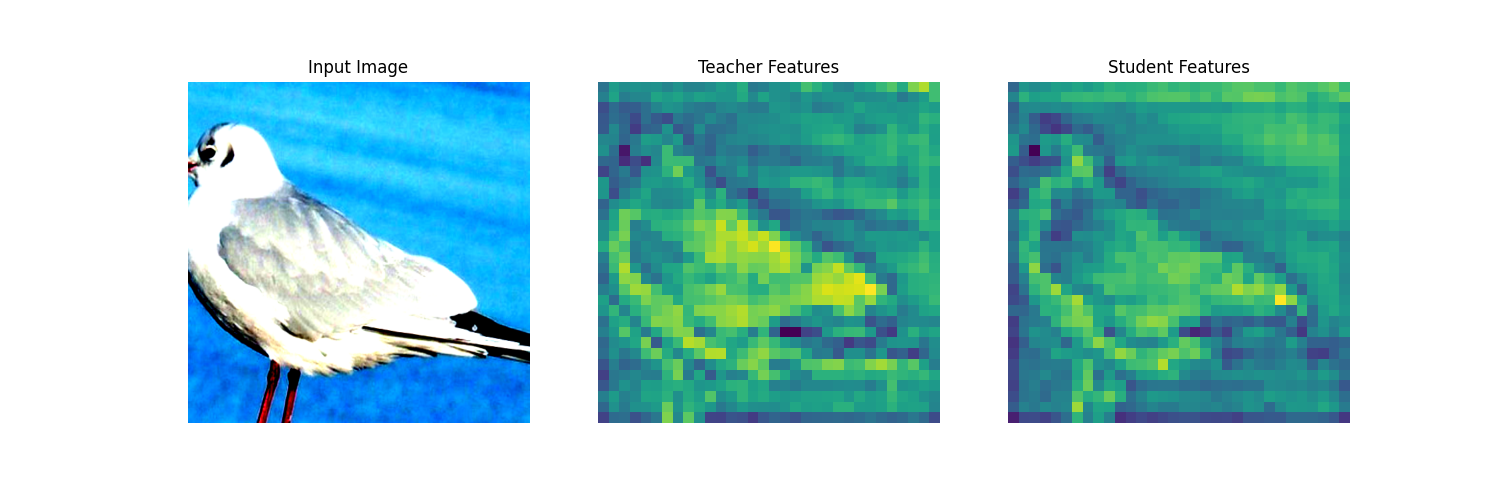}
  \end{subfigure}\hfill
  \begin{subfigure}{0.48\linewidth}
    \centering
    \includegraphics[width=\linewidth]{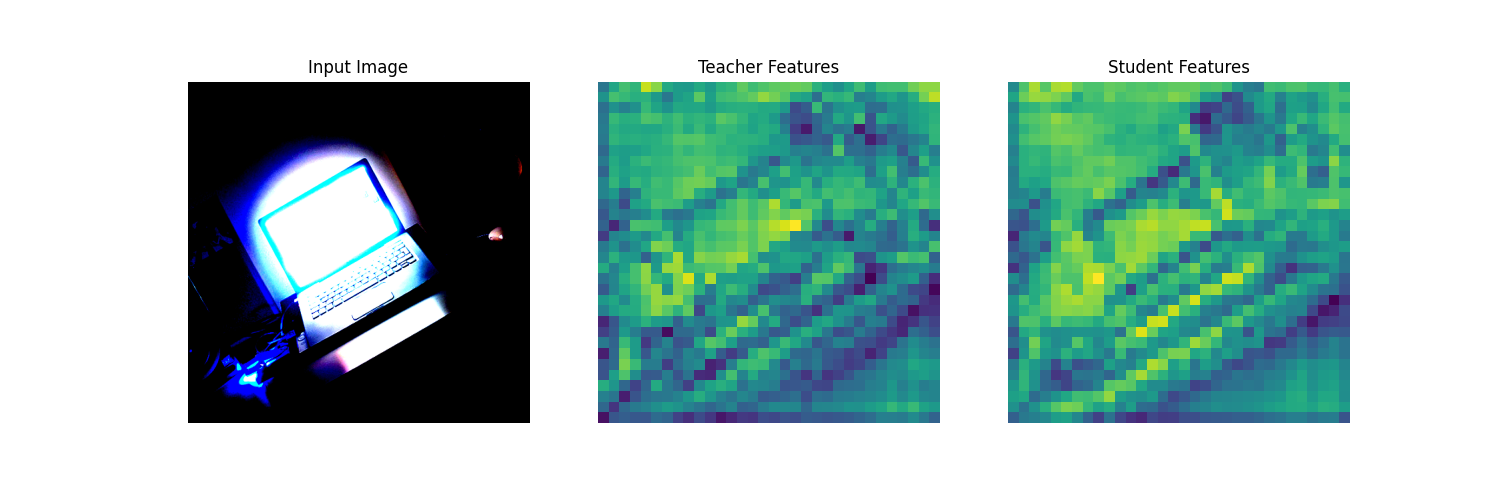}
  \end{subfigure}
  \caption{\textbf{NanoSAM2 distillation with Quant-Trim.} For two inputs (left in each subfigure), the student’s deepest FPN feature matches the teacher’s saliency structure while exhibiting fewer saturated patches after reverse pruning.}
  \label{fig:nanosam2-features}
\end{figure}

\begin{figure}[H]
    \centering
    \includegraphics[width=0.9\linewidth]{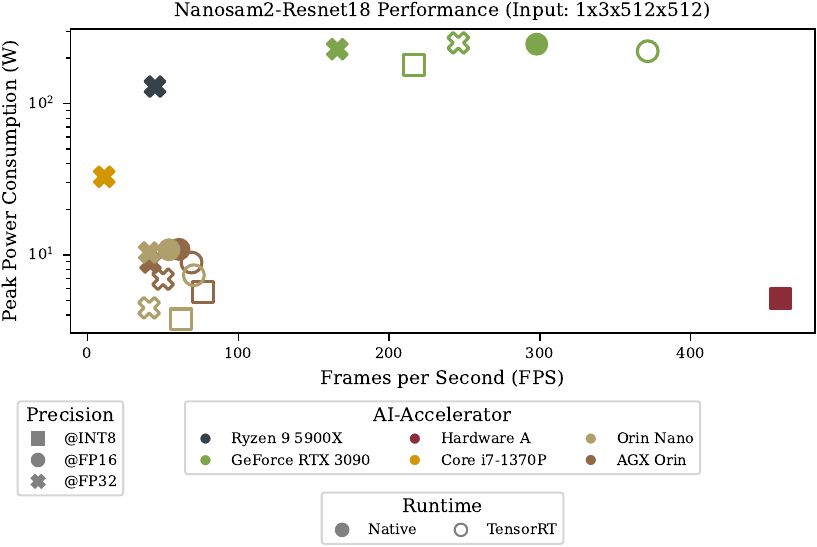}
    \caption{\textbf{End-to-end inference for NanoSAM2–ResNet-18 across accelerators.}
    Single-image, batch{=}1, $512{\times}512$ input; tiled inference with 50\% overlap when required. Results are averaged across runs and warmup runs are used for each hardware. Runtimes use TensorRT (FP16) on GPU/Jetson and vendor runtimes on NPUs (INT8 or BF16/INT as supported). 
    \emph{Hardware A}, which uses A8W8, is 6 times faster than the Jetson family hardware and faster than the desktop GPU at an FP16 baseline while operating at single-digit watts (about 5~W).}
    \label{fig:nanosam2-accel}
\end{figure}

We distill a compact NanoSAM2 image encoder from \emph{SAM-2.1 Hiera} inspired by the \texttt{nanosam} recipe\footnote{\nanosamrepo}, adapted to SAM-2.1 and our Quant-Trim curriculum (\cref{sec:quanttrim}). Distillation minimizes a three-scale FPN loss (Huber; weights $[1, \tfrac14, \tfrac18]$) between teacher and student features, while Quant-Trim runs on the student to align training numerics with INT deployment. NanoSam2 reaches an \textbf{mIOU} of \textbf{0.5889} on the Coco 20217 validation set \citep{lin2014coco}. Fig.~\ref{fig:nanosam2-features} shows representative cases: the student reproduces the teacher’s saliency structure (object contours, part boundaries) without high-frequency artifacts, and reverse pruning suppresses rare saturated responses that otherwise inflate activation ranges. Qualitatively, the feature maps remain sharp at object edges and smooth in background regions, indicating that Quant-Trim preserves the inductive content required for mask decoding while preparing the checkpoint for NPU compilers.
As shown in \cref{fig:nanosam2-accel}, NanoSAM2–ResNet-18 reaches real-time latencies on specialized NPUs at single-digit watts, surpassing a desktop FP16 GPU baseline in our setting. \emph{Implementation note:} the image encoder (dominant compute) runs on the NPU, while the lightweight prompt decoder runs on the host CPU.

\subsection{On-Device Deployment}
\begin{table}[H]
\centering
\begin{tabular}{lccccc}
\toprule
Method & Top-1 & Top-5 & MSE$\downarrow$ & Brier$\downarrow$ & ECE$\downarrow$ \\
\midrule
Quant-Trim & 64.60 (64.60) & 86.80 (86.80) & 0.01344 & 0.55235 (0.55070) & 0.19821 (0.19966) \\
MAP        & 64.30 (62.30) & 86.60 (85.80) & 0.03987 & 0.57446 (0.57182) & 0.22984 (0.23226) \\
\bottomrule
\end{tabular}
\caption{ResNet-50 on CIFAR-100, \mx  (INT8 weights, BF16 activations). Over 500 frames \textbf{average FPS}: 834.18 and  average System Latency: \textbf{5.34 ms}.
Entries are \textbf{On-Device}; FP32 values are in parentheses. 
Quant-Trim improves Top-1 by +2.3 pts and cuts MSE by $\sim$66\% vs.\ MAP, with near FP32–device parity across metrics.}
\label{tab:CIFAR-100_mx3_compact}
\end{table}

\begin{table}
\centering
\begin{tabular}{lccccc}
\toprule
Method & Top-1 & Top-5 & MSE$\downarrow$ & Brier$\downarrow$ & ECE$\downarrow$ \\
\midrule
Quant-Trim  & 65.10\ (64.60) & 86.90\ (86.80) & 0.01601 & 0.54295\ (0.55070) & 0.19791\ (0.19966) \\
MAP              & 62.70\ (62.30) & 85.70\ (85.80) & 0.02110 & 0.56826\ (0.57182) & 0.22860\ (0.23226) \\
\bottomrule
\end{tabular}
\caption{ResNet-50 on CIFAR-100, \sakura. Entries are \textbf{On-Device}; ONNX (FP32) references are in parentheses. MSE is computed as the mean squared difference between device and ONNX \emph{logits} (pre-softmax). \textbf{Average FPS}: 571, \textbf{IP execution time}: 1.5 ms}
\label{tab:CIFAR100_sakura_compact}
\end{table}

\paragraph{Beyond accuracy (MSE and calibration).}
We quantify backend drift using the mean squared error (MSE) between \emph{logits} (pre-softmax scores) produced on-device and by the ONNX FP32 reference: $\mathrm{MSE}=\tfrac{1}{N}\sum_i \lVert \text{device\_logits}_i-\text{onnx\_logits}_i\rVert_2^2$. On \mx, Quant-Trim cuts logit MSE from $0.03987$ (MAP) to $0.01344$ ($\downarrow\!\sim66\%$), with improved calibration (Brier $0.57446\!\to\!0.55235$, $\downarrow\!\sim3.8\%$; ECE $0.22984\!\to\!0.19821$, $\downarrow\!\sim13.7\%$) and a small Top-1 gain ($+0.3$ pts), see \cref{tab:CIFAR-100_mx3_compact}. On \sakura, Quant-Trim similarly reduces logit MSE from $0.021$ to $0.016$ ($\downarrow\!\sim24.2\%$) and improves calibration (Brier $0.56826\!\to\!0.54295$, $\downarrow\!\sim4.5\%$; ECE $0.22860\!\to\!0.19791$, $\downarrow\!\sim13.4\%$) while adding $+2.4$ Top-1 pts; see \cref{tab:CIFAR100_sakura_compact}. In both cases, lower MSE indicates closer numerical agreement to the FP32 reference at the \emph{logit} level (before probabilities are formed), and lower Brier/ECE reflect better probability calibration. For ResNets used as a \emph{feature backbone}, the lower MSE also signals higher backbone \emph{signal fidelity}, which helps keep the encoder usable for downstream heads and distillation targets—especially relevant on the edge where ResNet backbones commonly distil or interface with Vision Transformer (ViT) backbones in architectures like DETR, SAM and CLIP \citep{dosovitskiy2020vit,carion2020detr,kirillov2023sam,radford2021clip}. 

\paragraph{Signal-to-Noise Ratio (SNR)}
We show in \cref{tab:snr_comparison} that Quant-Trim demonstrates superior signal fidelity during deployment on \hailo, achieving an SNR of 43.12 on the output layer, even with only calibration enabled. In contrast, models trained using MAP, augmented with hardware-based optimization techniques like Equalization and Adaround (which require additional overhead), achieve a lower SNR of 34.3. This highlights Quant-Trim's ability to maintain higher signal quality without the need for extensive post-training adjustments, ensuring robust deployment across hardware platforms

\begin{table}[H]
\centering
\resizebox{\linewidth}{!}{
\begin{tabular}{lcc}
\toprule
\textbf{Training Method} & \textbf{SNR (Output Layer)} & \textbf{Training Details} \\
\midrule
Quant-Trim (Calibration Only) & 43.12 & no additional fine-tuning \\
Baseline (Equalization + Adaround \citep{nagel2020adaround}) & 34.30 & 320 epochs, 256 images, all layers quantized \\
\bottomrule
\end{tabular}
}
\caption{Comparison of Signal-to-Noise Ratio (SNR) on the output layer of ResNet-50 for Hardware A (A8W8 INT). Quant-Trim achieves significantly higher SNR with only calibration enabled compared to the baseline trained with Equalization + Adaround over 320 epochs.}
\label{tab:snr_comparison}
\end{table}

\section{Conclusion}
We presented \textit{Quant-Trim}, a training-time procedure that couples progressive fake quantization with reverse pruning to produce a single, hardware-agnostic checkpoint. Across CNNs and transformers, Quant-Trim narrows the FP$\rightarrow$INT gap, stabilizes training despite activation outliers, and reduces sensitivity to backend scaling, clipping, and operator coverage. On edge accelerators, the same checkpoint compiles reliably and delivers favorable latency/energy trade-offs without per-backend retraining or concrete graph edits. We further demonstrated practicality with NanoSAM2, showing on-device throughput gains while retaining accuracy. Future work will extend the evaluation to larger datasets and models.

\paragraph{Limitations} This work highlights a promising approach to optimizing neural networks for post-training quantization from the compiler's perspective. It is important to note that this is a work in progress. One challenge we are addressing separately is the deployability of these networks on specific hardware, as some architectures may not be suitable yet for deployment. The vendors and hardware that we tested are initially designed for particular architectures. Vendors are actively working to improve support for transformer architectures. Some vendors have either released custom hardware primarily suited for generative AI workloads, others are adjusting their compilers for enhanced compatibility, showing the promise of Edge Deployment.
Our goal is not to compare one device to another but rather improve the deployment results for each hardware as selecting a specific NPU for an application highly depends on the task at hand, operation support and the size of the graph. We are committed to advancing efficient green AI solutions. While our approach is primarily designed to meet stringent real-world application requirements, it also scales up to GPU levels. With further tweaking and optimization, we can reduce the footprint of larger GPUs. Future evaluations on larger datasets will enhance this work, as we are currently addressing both training and inference; the experiment load is massive, thus we will include a broader evaluation in a future version.

\newpage

\bibliographystyle{abbrv}
\bibliography{references} 

\newpage

\appendix

\section{Experimental Setup}\label{sec:experm}

\subsection{Hardware Platforms}\label{sub-sec:hardplatform}

The performance and accuracy of a quantized model are fundamentally tied to the capabilities of the target hardware, including its supported precisions, compiler toolchain, and supported operations. To evaluate the robustness of our method, we benchmark across two distinct classes of edge devices: versatile embedded GPUs and specialized AI accelerators (NPUs).

\begin{itemize}
    \item \textbf{Embedded GPUs:} We use NVIDIA's Jetson Nano and Jetson AGX Orin. These platforms offer general-purpose CUDA and Tensor Core acceleration with broad support for FP32, FP16, and INT8. Their compilers (e.g., TensorRT) provide significant flexibility, including support for \textbf{dynamic quantization}, where activation scales are computed on-the-fly. This adapts to varying input data distributions but can introduce runtime overhead.
    \item \textbf{Specialized NPUs:} We benchmark on three representative edge AI accelerators anonymized as Hardware A, B, and C and provide their specifications. These ASICs are optimized for high-throughput, low-power inference, primarily supporting aggressive INT8 (even INT4) quantization, while it is possible to set some layers to INT16; that might risk not fitting into the memory. Their toolchains are often more restrictive and typically rely on \textbf{static quantization}, where activation ranges are pre-calibrated offline using a representative dataset. This minimizes runtime computation but makes model performance highly sensitive to the calibration data and potential distribution shifts during deployment.
\end{itemize}

\begin{table}[h]
\centering
\resizebox{\linewidth}{!}{
\begin{tabular}{l l l l c}
\toprule
\textbf{Device} & \textbf{Type} & \textbf{W/A path} & \textbf{Act.\ scaling @ inference} & \textbf{PTQ calib.\ (INT)} \\
\midrule
Jetson Nano       & SoC (GPU)      & \WA{8}{8} \;or\; \WA{8}{16} & \stat{} (INT) or \qat{}       & \cond  \\
Jetson AGX Orin   & SoC (GPU)      & \WA{8}{8}, \WA{8}{16}       & \stat{} (INT) or \qat{}       & \cond  \\
\hailo            & M.2 NPU        & \WA{8}{8} (INT8)            & \stat{} (no runtime dyn)      & \yes \\
\sakura           & M.2/PCIe NPU   & \WA{8}{8} (INT8) or BF16    & \stat{} (compiler-provided)   & \no \\
RK3588 (RKNN)     & SoC (NPU)      & \WA{8}{8} \;or\; FP16       & \stat{} (INT8 only)           & \cond \\
\mx               & M.2 NPU        & \WA{8}{BF16} (hybrid)       & \na{} (A=BF16)                & \no \\
\bottomrule
\end{tabular}
}
\vspace{2pt}
\caption{\textbf{Quantization behavior (corrected).} \emph{W/A path} denotes weight/activation precisions at inference. 
\emph{Act.\ scaling}: \stat{} = fixed/static activation ranges used by the backend (may come from compiler defaults or embedded QAT scales); \qat{} = scales learned during QAT and embedded in the graph. 
\emph{PTQ calib.\ (INT)} indicates whether a representative dataset is required \emph{when an INT mode is selected}. 
For RK3588, calibration is \cond{}: required for INT8, not for FP16 or BF16.}
\label{tab:hardware-quant-compact}
\end{table}

\subsection{Hardware Form Factors and Practical Advantages}
\label{subsec:form-factors}
Edge accelerators appear in two dominant forms:
\begin{itemize}
    \item \textbf{Add-in NPUs }(M.2 / PCIe, USB).
Examples include \hailo (M.2 B/M-key), \mx (M.2), and  \sakura (M.2/PCIe variants). These modules attach to a host (x86/ARM SBC, laptop, or embedded carrier) over PCIe (or occasionally USB). The host handles I/O, pre/post-processing, and scheduling; the NPU executes the neural network graph. Advantages: \emph{drop-in} acceleration for existing systems, low incremental power (single-digit watts), and flexible host software. Caveats: host–device transfers (DMA), graph partitioning, and op coverage can introduce tail latency if unsupported ops fall back to the host.

\item \textbf{System-on-Chip} (SoC) platforms.
Examples include NVIDIA Jetson (Orin family) and Rockchip RK3588 boards. These integrate CPU, GPU/NPU, DRAM, and peripherals on one module; the full application stack (capture, pre/post, model, control) runs on-device. Advantages: single box, deterministic I/O, simpler memory topology, and easy CPU fallback for unsupported ops. Caveats: total power headroom is fixed; thermal throttling and shared memory bandwidth must be managed.
\end{itemize}

\begin{table}
\centering
\resizebox{1\linewidth}{!}{
\begin{tabular}{lccc}
\toprule
\textbf{Form factor} & \textbf{Typical link} & \textbf{Strengths} & \textbf{Watch-outs} \\
\midrule
M.2 / PCIe NPU & PCIe Gen3/4 x2–x4 & Drop-in accel; low watts; scalable & PCIe copies; op coverage; host fallbacks \\
USB NPU stick  & USB 3.x            & Quick prototyping; portable       & Higher copy overhead; limited bandwidth \\
SoC (Jetson)    & On-package MMU/DRAM & Unified memory; full stack on device & Thermal/power budgets; shared BW \\
SoC (RK3588)    & On-package MMU/DRAM & Low cost; rich I/O                & Compiler maturity; INT-centric kernels \\
\bottomrule
\end{tabular}
}
\caption{Form factors, data links, and common trade-offs at the edge.}
\label{tab:formfactor}
\end{table}

\begin{table}
\centering
\footnotesize
\begin{tabular}{l l c c}
\toprule
\textbf{Accelerator} & \textbf{Form factor / Interface} & \textbf{Peak perf.} & \textbf{Typical power} \\
\midrule
\hailo${}^{\dagger}$   & M.2 2280 (B/M); PCIe Gen3 x2 & 26\,TOPS (INT8) & $\sim$2.5\,W \\
\mx${}^{\ddagger}$     & Chip (PCIe Gen3 x4 / USB3); M.2 module & 6\,TOPS/chip (INT8) & 0.5–2\,W/chip \\
\sakura${}^{\S}$       & Low-profile PCIe; Gen3 x8    & 60\,TOPS (INT8) & 8–10\,W \\
\bottomrule
\end{tabular}
\caption{\textbf{Edge NPU summary (TOPS \& typical power).}
Vendor-quoted peak performance; INT8 unless noted. Symbols: ${}^{\dagger}$ \hailo uses on-chip SRAM only (no external DRAM); ${}^{\ddagger}$ \mx numbers are per chip (M.2 module aggregates \emph{4} chips, up to $\sim$24\,TOPS); ${}^{\S}$ \sakura  also reports $\sim$30\,TFLOPS (BF16), typical 8–10\,W.}%
\label{tab:edge-npu-specs}
\end{table}

\paragraph{Operator support and fallbacks.}
Add-in NPUs excel when the compiled graph maps \emph{entirely} to accelerator kernels. If a layer is unsupported (or precision constraints force a dequant–requant island), the compiler will in some cases route that subgraph to the host CPU/GPU; it is also possible to crop and run only a subgraph on the NPU if the issue is only in the pre- or post-processing layers. This is functionally correct but may introduce large latency spikes due to extra memory traffic and synchronization. SoCs tolerate such cases better (same memory space, no PCIe), though overall throughput may be lower.

\paragraph{Pre/post-processing placement.}
Lightweight steps (resize, normalization, tiling) should run where they minimize copies: on SoCs, prefer GPU/NPU-side ops with zero-copy buffers; on M.2 NPUs, batch/pack on the host, transfer once, and fuse post-ops into the compiled graph where supported. In both cases, avoid alternating host–device–host hops in the critical path.

\paragraph{Memory and bandwidth.}
Edge NPUs rely on on-chip SRAM and tiling to maintain high reuse; DRAM is accessed in bursts. Sustained performance is determined as much by \emph{dataflow} (blocking/tiling, operator fusion, quantized layout) as by peak TOPS. SoCs share DRAM across CPU/GPU/NPU, so contention and cache policy can affect tail latency; pinning cores and using contiguous DMA buffers often helps.

\subsection{Appendix: Hyperparameters and Metrics}
\label{app:hyperparams}
Reverse pruning: every $K$ epochs, clip weights to the $p_{\text{clip}}$ percentile (per-tensor or per-channel). Running ranges use EMA with momentum $\mu$.
\paragraph{Measurement protocol.}
Inference: \emph{20 warmup + 200 timed} iterations; medians over \textbf{5 runs}. Training: medians over \textbf{3} random seeds.

\begin{table}
\centering

\caption{Minimal QAT defaults by task/dataset.}
\label{tab:qat_defaults}

\begin{tabular}{lcc}
\toprule
 & \textbf{CIFAR-100 } & \textbf{Segm. (COCO)} \\
\midrule
Epochs / Batch & 100 / 128 & 100 / 32 \\
Optimizer / LR & AdamW / $3{\times}10^{-4}$ & AdamW / $5{\times}10^{-4}$ \\
Weight decay & 0.01 & $1{\times}10^{-4}$ \\
LR schedule & Cosine & Cosine \\
$E_w, E_f, H$ & 10, 50, 20 & 15, 30, 20 \\
$p_{\text{clip}}$ / $K$ & 0.90 / 5 & 0.95 / 5 \\
EMA $\mu$ & $10^{-2}$ & $10^{-3}$ \\
Target precision & INT8 (W/A) & INT8 (W/A) \\
\bottomrule
\end{tabular}
\end{table}

\vspace{-4pt}
\begin{table}[H]
\centering
\small
\caption{Architecture-specific tweaks.}
\label{tab:model_tweaks}
\resizebox{1\linewidth}{!}{
\begin{tabular}{lcc}
\toprule
 & \textbf{ResNet (CNN)} & \textbf{DINOv2 (Transformer)} \\
\midrule
LR / warmup & higher LR; shorter $E_w$ (10--30) & lower LR; longer $E_w$ (30--50) \\
Ramp length $E_f$ & 30--50 & 60--100 \\
$p_{\text{clip}}$ / $K$ & 0.90--0.95 / 5 & 0.96--0.98 / 15 \\
EMA $\mu$ & $10^{-3}$--$10^{-2}$ & $10^{-4}$--$10^{-3}$ \\
Attention handling & n/a & Q/K/V and outputs fake-quant; keep scores in FP \\
Final blend cap & $\alpha_{\max}{=}1.0$ & $\alpha_{\max}{\approx}0.8$ \\
\bottomrule
\end{tabular}}
\end{table}

\vspace{2pt}
\paragraph{Metrics (reported).}
\textbf{Classification}: Top-1/Top-5. \;
\textbf{Segmentation}: mean Intersection over Union (mIoU), with $\text{mIoU} = \frac{1}{N} \sum_{i=1}^{N} \text{IoU}_i = \frac{1}{N} \sum_{i=1}^{N} \frac{|A_i \cap B_i|}{|A_i \cup B_i|}$
\textbf{Calibration/robustness}: ECE, MSE (logits). \;
\textbf{Efficiency}: FPS, average power (W).

\subsection{Models}

To ensure robustness across architectures, we evaluate:
\begin{itemize}
    \item \textbf{ResNet-50 and ResNet-18}~\cite{he2016resnet}: canonical CNN backbone with residual connections.
    \item \textbf{NanoSAM2}: SAM variant with a ResNet-18 backbone; we use knowledge distillation to train the model on COCO.
    \item \textbf{DINOv2}~\cite{oquab2023DINOv2}: self-supervised vision transformer trained on large-scale data, challenging to quantize.
\end{itemize}

\subsection{Datasets}

We evaluate on:
\begin{itemize}
    \item \textbf{CIFAR-100}~\cite{krizhevsky2009cifar}: 100-class dataset of tiny natural images (32$\times$32). Serves as a proxy for classification robustness under quantization.
    \item \textbf{MS-COCO}~\cite{lin2014coco}: large-scale benchmark for detection and segmentation. Stress-tests activation quantization due to varied input scales and long-tail distribution.
      \item \textbf{CIFAR-10}~\citep{krizhevsky2009cifar}: Used for ablations due to its lightweight compute footprint, enabling rapid sweeps over clipping percentiles and fake-quant schedules.
\end{itemize}

\section{Ablation Study: Quantization Components and Clipping Sensitivity}\label{sec:ablation}

We conduct a systematic ablation study on ResNet-18 with CIFAR-10 to isolate the contributions of different quantization components in our framework. The study evaluates five experimental configurations to understand the individual and combined effects of fake quantization training  and reverse pruning (outlier weight removal). The experimental setup is described in \cref{tab:ablation-setup}. The five configurations isolate the role of our fake-quantization and reverse pruning ; all share identical optimizer and schedule.

\begin{table}[h]
\centering
\small
\resizebox{\textwidth}{!}{%
\begin{tabular}{lcccc@{\hspace{8pt}}l}
\toprule
\textbf{Config} & \textbf{Fake-Quant (INT8)} & \textbf{Reverse-Pruning} & $\mathbf{p_{\text{clip}}}$ & \textbf{Epochs} & \textbf{Notes} \\
\midrule
(1) FP32 Baseline            & \xmark  & \xmark & --    & 50 & Standard full-precision training \\
(2) QAT Only                 & \cmark  & \xmark & --    & 50 & INT8 fake-quantization; RP disabled \\
(3) Reverse Pruning Only     & \xmark  & \cmark & 95\%  & 50 & FP32 training with percentile clipping \\
(4) QAT + 90\% Clipping      & \cmark  & \cmark & 90\%  & 50 & Aggressive outlier removal \\
(5) QAT + 99\% Clipping      & \cmark  & \cmark & 99\%  & 50 & Conservative outlier removal \\
\bottomrule
\end{tabular}
}
\vspace{4pt}
\caption{\textbf{Ablation configurations} for ResNet-18 on CIFAR-10. Shared hyperparameters across all runs: SGD optimizer, learning rate $10^{-3}$, weight decay $5{\times}10^{-4}$, $50$ epochs, and $n{=}3$ seeds. Only quantization settings differ between rows.}
\label{tab:ablation-setup}
\end{table}
 
\begin{figure}[t]
    \centering
    \includegraphics[width=\linewidth]{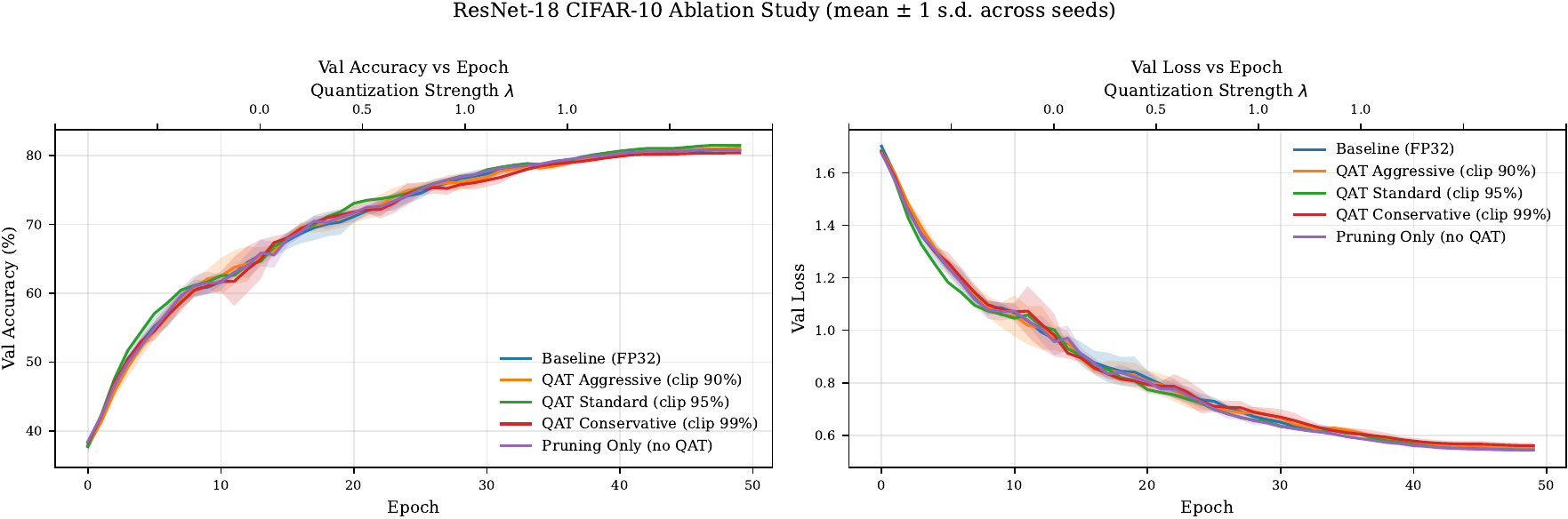}
    \caption{\textbf{Ablation study on ResNet-18 CIFAR-10 shows convergence to similar accuracy across configurations.} 
    \textbf{(Left)} Validation accuracy versus training epoch. Despite varying quantization strategies (aggressive clip 90\%, standard clip 95\%, conservative clip 99\%) and the baseline FP32 model, all configurations converge to approximately 81\% validation accuracy by epoch 50, demonstrating that the quantization strength schedule $\lambda$ (top axis) does not significantly impact final model performance. 
    \textbf{(Right)} Validation loss follows a consistent downward trend across all methods, further confirming stable convergence behavior. 
    The pruning-only baseline (no QAT) achieves comparable performance, suggesting that the learned representations are robust to different compression approaches. 
    Shaded regions indicate $\pm 1$ standard deviation across 3 random seeds per configuration.}
    \label{fig:ablation_convergence}
\end{figure}

\begin{figure}
\centering
\includegraphics[width=0.49\textwidth]{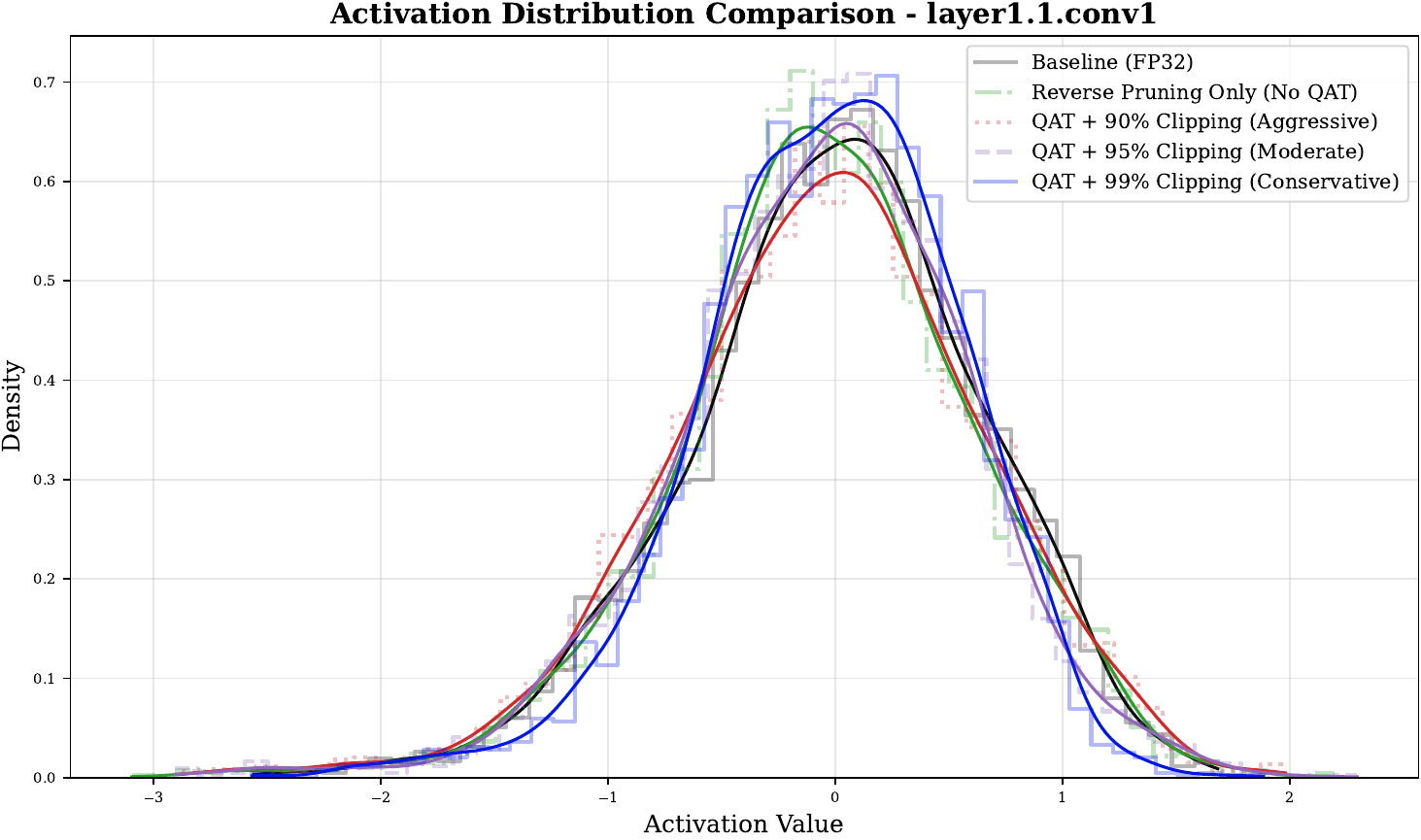}
\includegraphics[width=0.49\textwidth]{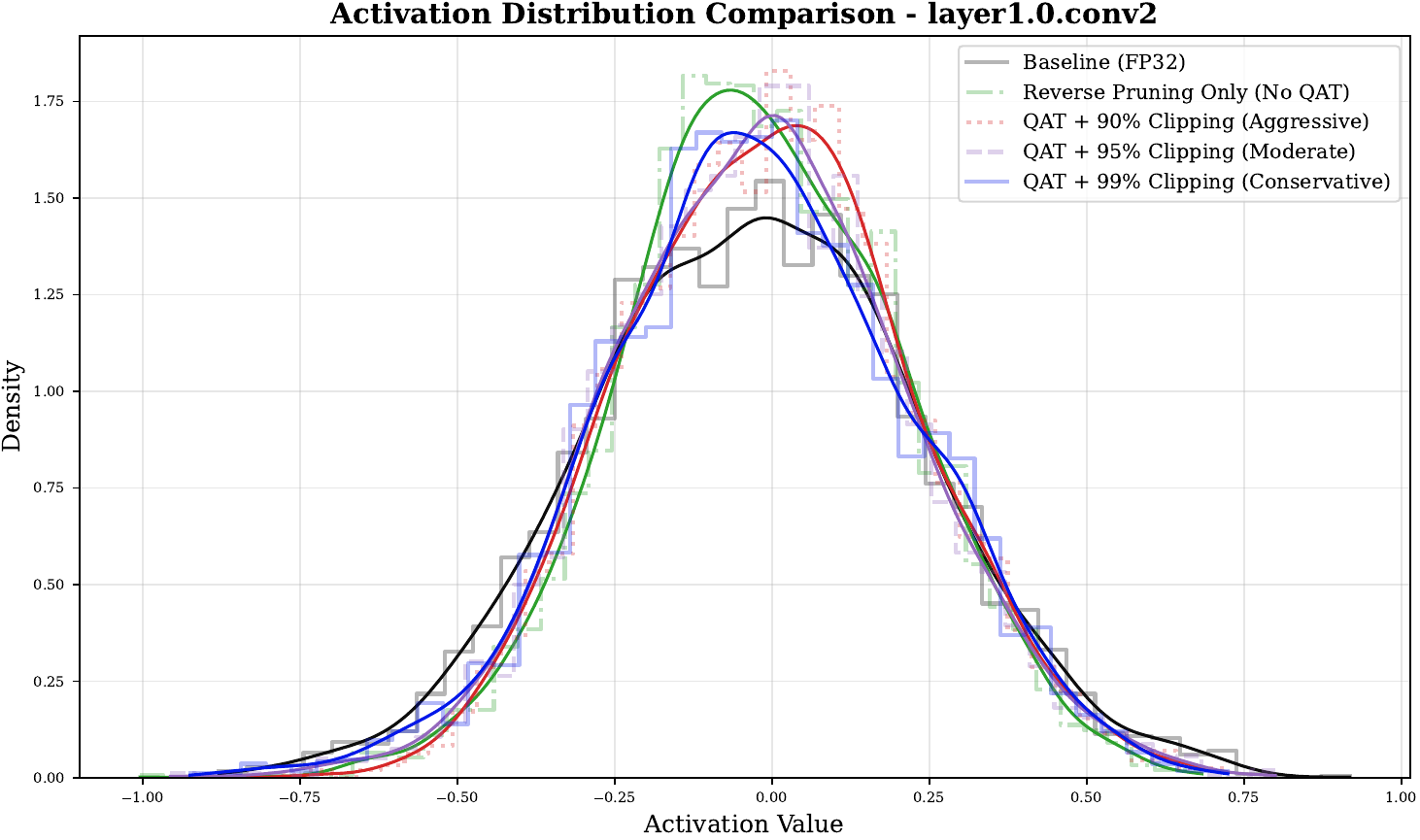}
\caption{Weight distribution comparison across ablation study configurations for ResNet-18 on CIFAR-10. The baseline FP32 model (black) exhibits the widest weight distribution, while quantization-aware training introduces characteristic distribution narrowing. Reverse pruning alone (green dash-dot) provides regularization effects visible as distribution tightening around zero. The combination of QAT with different clipping percentiles reveals the trade-off between outlier removal and weight preservation: aggressive 90\% clipping (red dotted) creates the most \textbf{constrained} distribution, while conservative 99\% clipping (blue) maintains broader weight ranges and \textbf{95\% }being the sweet spot (purple) where it is observed in a \textbf{low MSE of 0.00023} between compiled model and FP32 predictions on \mx.}
\label{fig:ablation_weight_distributions}
\end{figure}

Our experiment reveals that on the training level, all ablation configurations converge to similar validation accuracy ($\approx$81\%) despite varying quantization strategies and clipping thresholds (90\%, 95\%, 99\%) as seen in \cref{fig:ablation_convergence}.
The convergence of both the FP32 baseline and pruning-only variant to comparable performance levels demonstrates that our progressive quantization strength schedule $\lambda$ effectively maintains model capacity across diverse compression approaches.
This robustness to hyperparameter choices validates the stability of the quantization-aware training framework and suggests that the specific clipping threshold has minimal impact on final model performance. In addition in \cref{fig:ablation_weight_distributions}, We can see based on the activation distribution that for our QAT approach paired with reverse pruning at  95\% results in a smoother activation distribution that is easier to quantize and with less outliers.

\section{Pseudo code}

\begin{algorithm}[H]
\caption{Quant-Trim Training}
\label{alg:quanttrim}
\begin{algorithmic}[1]
\STATE Initialize weights $w^{(0)}$ in FP32
\FOR{epoch $t=1,\dots,T$}
    \IF{$t=E_w$}
        \STATE Reverse prune (all $\ell$): $w_\ell \leftarrow \mathrm{clip}(w_\ell,-\tau_{\ell,t},\tau_{\ell,t})$ with $\tau_{\ell,t}=(1-\beta)\tau_{\ell,t-1}+\beta\,\widehat Q^{(S)}_{|w_\ell|}(p_{\mathrm{clip}})$
    \ENDIF
    \STATE Update robust stats $(s_t^{(\mathrm{w})},z^{(\mathrm{w})}{=}0)$ and $(s_t^{(\mathrm{a})},z_t^{(\mathrm{a})})$ via EMA percentiles
    \STATE Set $\lambda_t$ by the schedule above
    \STATE Fake-quant forward: $\hat{w}=Q_b(w;s_t^{(\mathrm{w})},0)$, $\hat{x}=Q_b(x;s_t^{(\mathrm{a})},z_t^{(\mathrm{a})})$, output $\tilde{x}=x+\lambda_t(\hat{x}-x)_{\text{stop-grad}}$
    \STATE Backprop with STE on FP32 master weights
\ENDFOR
\STATE Export checkpoint $\to$ ONNX (compile with TensorRT/TVM/NPU compilers)
\end{algorithmic}
\end{algorithm}

\section{Additional Results}

\begin{figure}
    \centering
    \includegraphics[width=1\linewidth]{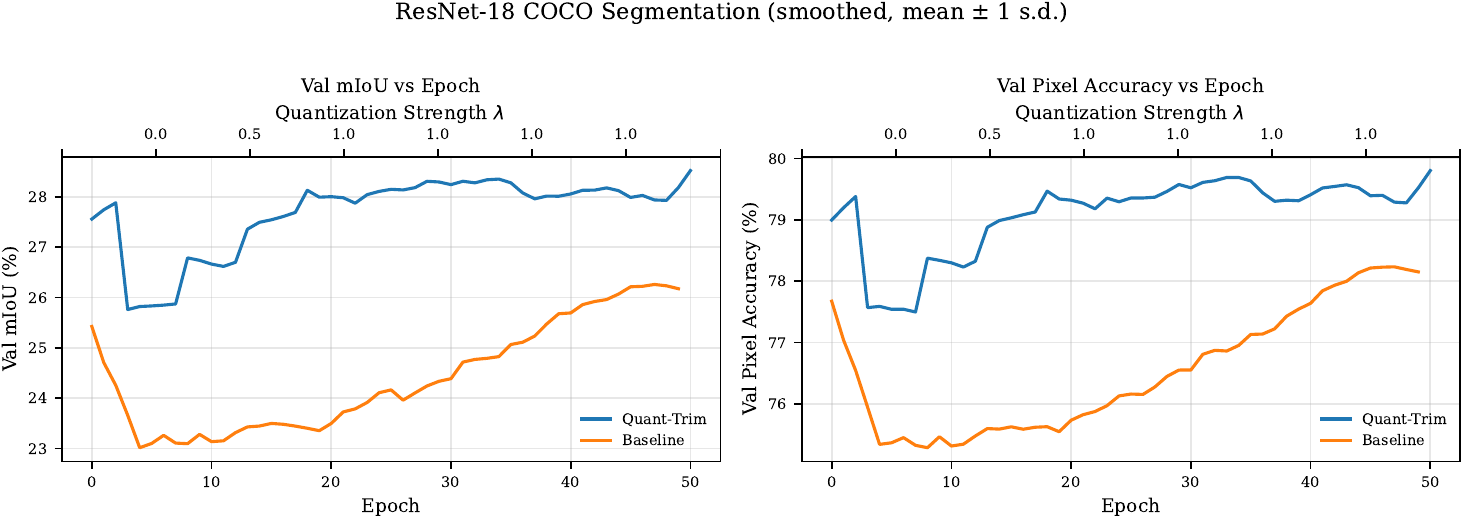}
    \caption{ResNet-18 COCO Segmentation: Val mIoU and Pixel Accuracy vs Epoch}
    \label{fig:resnet_coco}
\end{figure}

\begin{table}[H]
\centering
\label{tab:hardware_nano}
\resizebox{\textwidth}{!}{%
\begin{tabular}{l l c c c c c}
\toprule
Hardware & Type & Price & Peak Power (W) & Runtime env (ACC) & Runtime (s) & Price per Watt (€) \\
\midrule
RTX 3090 (comparison) & GPU & 1500€ & 190 & TensorRT (FP16) & 0.12 & 0.127 \\
Jetson Orin Nano 8 GB & SOM & ~250€ & 10 & TensorRT (FP16) & 0.66 & 0.040 \\
\hailo & M.2 Mod. & ~150€ & 5 & (INT8) & 0.10 & 0.033 \\
\mx & M.2 Mod. & ~125€ & 5 & (BF16) & 0.60 & 0.040 \\
\rknn & Full SoC & ~250€ & 8 & (INT/FP16) & 3.50 & 0.032 \\
\sakura & M.2 Mod. & ~350€ & 8 & (INT/FP16) & 0.76 & 0.023 \\
\bottomrule
\end{tabular}
}
\caption{NanoSAM2 backbone runtime for one 2k$\times$2k image (50 tiles). We report backbone latency only; the lightweight decoder runs on CPU. Images larger than 1024 px are processed by tiled inference (512$\times$512 tiles with 50\% overlap).}
\end{table}

\begin{figure}
\centering
\includegraphics[width=\textwidth]{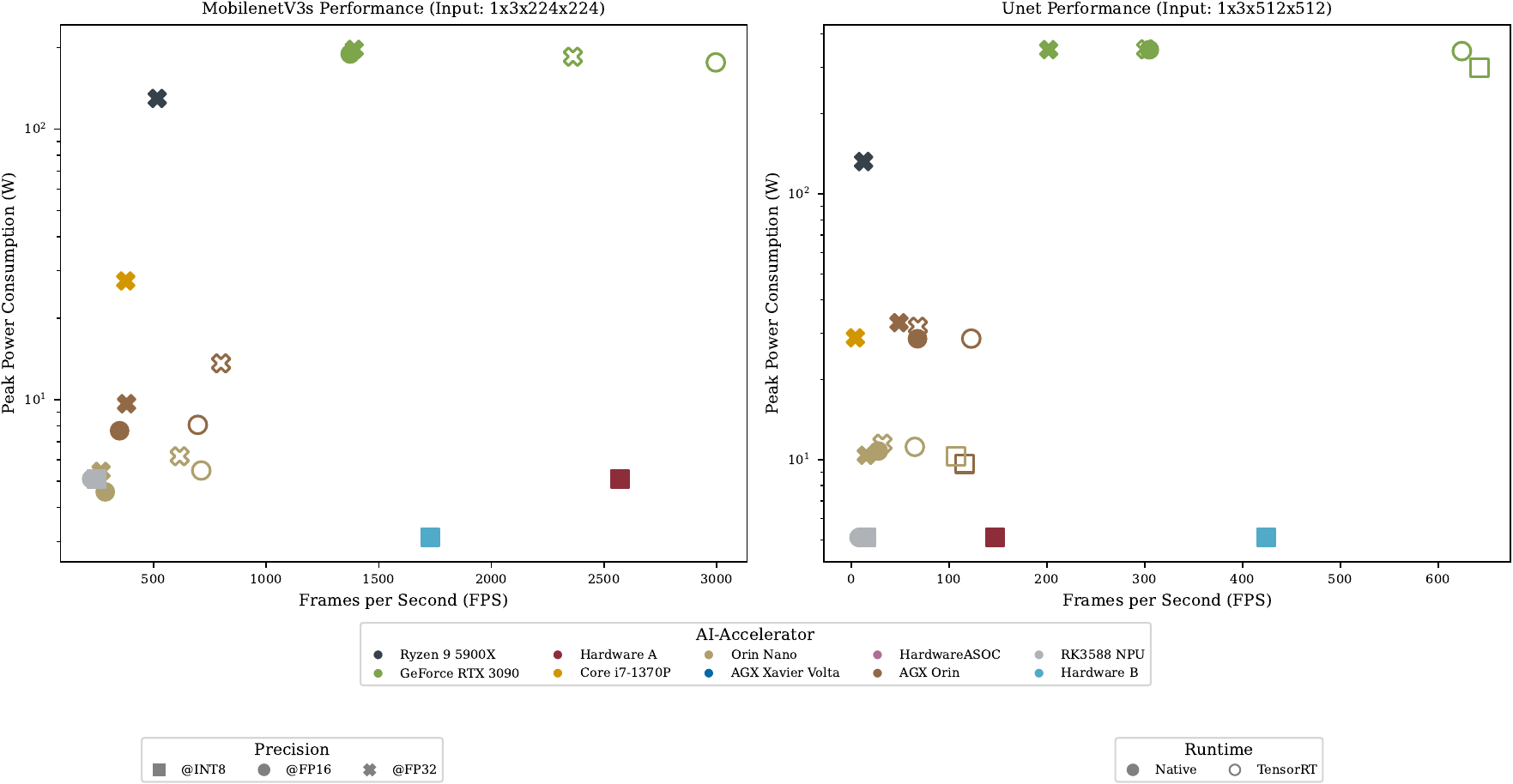}
\caption{Performance and power consumption comparison across different AI accelerators for MobileNetV3s and U-Net models. }
\label{fig:mobilenet_unet_comparison}
\end{figure}

\newpage




\end{document}